\documentclass[letterpaper,10pt,conference]{ieeeconf}

\IEEEoverridecommandlockouts
\usepackage{graphicx}
\usepackage{cite}
\usepackage{amsmath,amssymb,bm}
\usepackage{booktabs}
\usepackage{array}
\usepackage{balance}
\usepackage{placeins}

\graphicspath{{figures/}}

\newcommand{\missingfigure}[1]{%
  \fbox{\parbox[c][32mm][c]{0.94\linewidth}{\centering #1}}%
}
\newcommand{\includeorplaceholder}[3]{%
  \IfFileExists{#1.jpg}{%
    \includegraphics[width=#2]{#1.jpg}%
  }{%
    \IfFileExists{#1.png}{%
      \includegraphics[width=#2]{#1.png}%
    }{%
      \IfFileExists{#1.pdf}{%
        \includegraphics[width=#2]{#1.pdf}%
      }{%
        \missingfigure{#3}%
      }%
    }%
  }%
}

\title{\LARGE \bfseries
A Support-Enhanced Granular-Jamming Gripper for RL-based Grasping with Continuum Manipulators
}

\author{Danyu Liu$^{1,2}$,
        Tianlin Zhang$^{1,2}$,
        Wei Chen$^{1,2}$,
        Wei Tang$^{3}$,
        Kecheng Qin$^{1,2,*}$,
        and Zhongyu Li$^{1,2,*}$%
\thanks{$^{1}$Hong Kong Embodied AI Lab, Hong Kong SAR, China.}%
\thanks{$^{2}$The Chinese University of Hong Kong, Hong Kong SAR, China.}%
\thanks{$^{3}$Zhejiang University, Hangzhou, China.}%
\thanks{$^{*}$Corresponding authors:
\{zhongyuli, kechengqin\}@cuhk.edu.hk.}%
}

\begin{document}

\maketitle
\thispagestyle{empty}
\pagestyle{empty}

\begin{abstract}
Continuum manipulators provide dexterous motion in confined spaces, but structural compliance, hysteresis,
and load-dependent deformation leave residual position and
orientation errors that can undermine reliable contact with
rigid grippers. To address this limitation, this paper presents a
lightweight support-enhanced granular-jamming gripper tailored to a continuum manipulator. The gripper maintains compliance before
jamming while establishing a direct load path to the continuum manipulator tip after jamming. To improve its grasping
performance, we systematically designed membrane materials,
particles, filling ratios, and the internal support structure, and
further identify geometry-dependent grasp boundaries with
respect to contact offset and object shape. Building on these
results, we construct a physical manipulation system integrating the continuum manipulator, granular-jamming gripper,
visual feedback, tendon actuation, and pneumatic control. We
then train a reinforcement-learning-based reaching controller
in a randomized simulation and deploy it on the physical
system, demonstrating how positioning control and contact level mechanical adaptation can complement each other in a
modular grasp-and-release task. By introducing an adaptive
structure that relaxes the need for highly accurate modeling
and positioning control, this work explores a design paradigm
that integrates physical and embodied intelligence.
\end{abstract}

\section{Introduction}

Continuum manipulators generate motion through continuous deformation of a
compliant body. Unlike mechanisms composed of rigid links and discrete joints,
they are slender, conformable, and capable of following curved paths
~\cite{Webster2010ConstantCurvature}. These properties enable operation in
confined and unstructured environments where conventional rigid manipulators
may have limited access~\cite{BurgnerKahrs2015Medical}. As continuum robots
progress from inspection toward physical manipulation, reliable grasping and
object transfer become increasingly important.

The compliance that enables dexterous motion also makes the end-effector state
difficult to predict and reproduce. Tendon friction, backlash, material
nonlinearity, external loading, and body deformation affect the relationship
between tendon actuation and end-effector motion
~\cite{Rao2021ModelTDCR}. Visual feedback can reduce accumulated positioning
error by correcting motion from measured end-effector states
~\cite{Yip2014ModelLess}. However, finite position and orientation errors, as
well as small residual motions, can remain during physical operation.

\begin{figure}[!t]
  \centering
  \includeorplaceholder{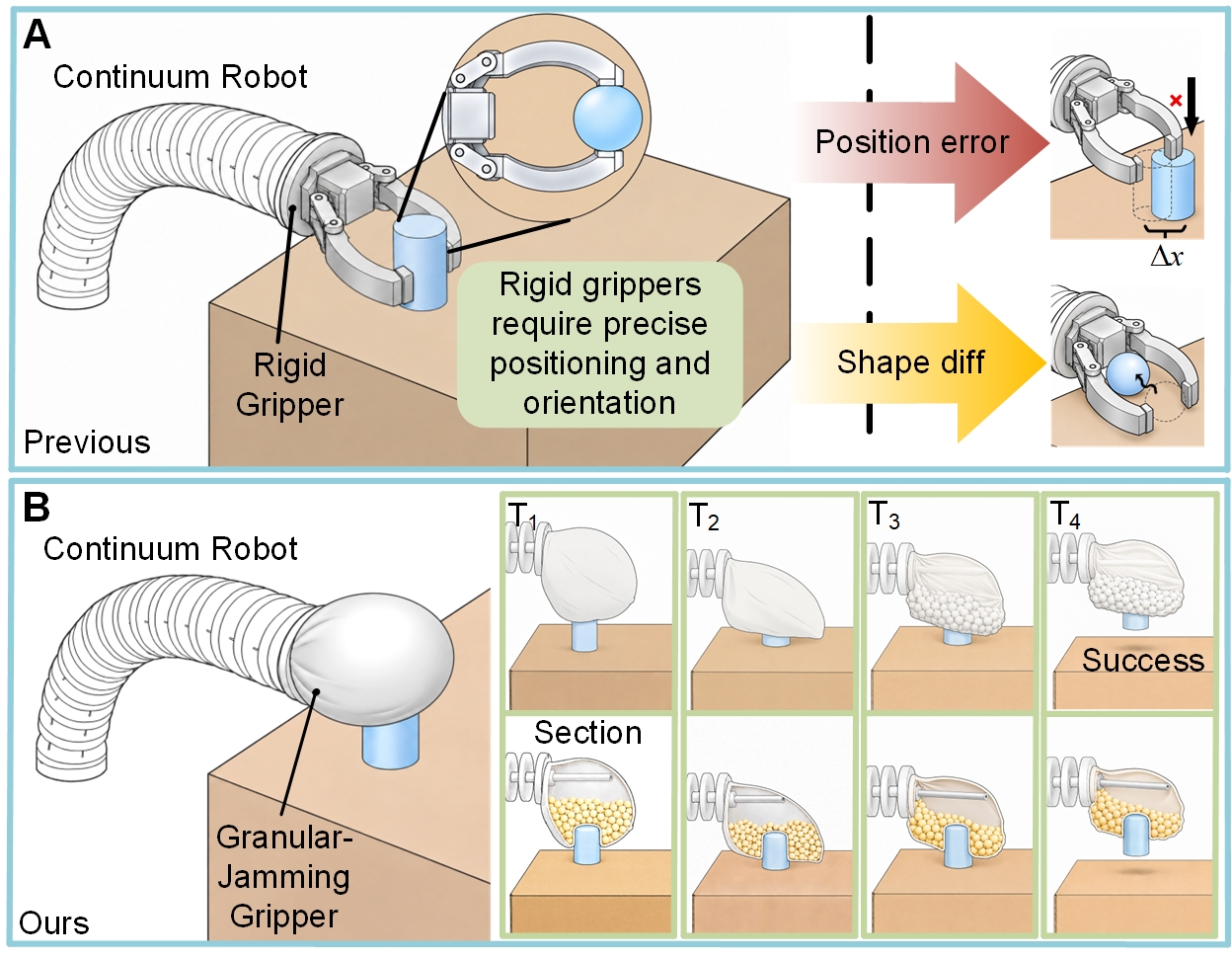}{0.98\columnwidth}{%
    Upload Fig. 1 as figures/fig1\_concept.jpg, .png, or .pdf.}
  \caption{Motivation for mechanically adaptive grasping with a continuum
    manipulator. (A) A rigid gripper requires sufficiently accurate relative
    position, orientation, and geometric compatibility; residual deviations can
    cause collision or prevent stable closure. (B) The unjammed granular gripper
    deforms through contact and continued compression to envelop the object,
    after which vacuum-induced jamming locks the adapted configuration. The
    comparison illustrates how contact-induced conformity relaxes the requirement
    for a single precise pre-contact pose.}
  \label{fig:concept}
\end{figure}

These residuals become especially important during grasping. A rigid gripper
requires a suitable relative pose and sufficient geometric compatibility
before closure. A small error at the continuum tip can therefore cause
gripper--object collision, loss of contact, or failure to establish a stable
constraint. Mechanically adaptive grippers provide an alternative by using
compliant deformation to accommodate uncertainty during contact
~\cite{Shintake2018SoftGrippers}. The challenge is to retain this adaptability
without introducing excessive distal mass, overall size, or actuation
complexity. The contrast between rigid and mechanically adaptive grasping is
illustrated in Fig.~\ref{fig:concept}.

Autonomous grasping introduces a further control challenge. For a vision-guided tendon-driven continuum manipulator, the uncertain response of the compliant body complicates the relationship between tendon actuation and observed end-effector motion. Learning-based control can
capture complex continuum behavior from data
~\cite{Thuruthel2017LearningDynamics}, but a policy trained in simulation must
still tolerate differences in posture, stiffness, actuation timing, and target
location when deployed on the physical system. The reaching controller must
therefore bring the gripper into a contactable region without assuming that it
can eliminate every contact-level error.

This work presents a lightweight, support-enhanced granular-jamming gripper
tailored to a tendon-driven continuum manipulator. The gripper remains
compliant before evacuation and conforms to the object through contact and
continued compression. An internal support rod promotes lateral enclosure and
transmits compressive force during this process. After jamming, the same
structure provides a more direct load path from the grasped object to the
continuum tip. We characterize the gripper configuration and its
geometry-dependent grasp boundaries, and then integrate it with visual
feedback, reinforcement-learning-based reaching, tendon actuation, and
pneumatic task control to demonstrate autonomous grasping and release.

The \textit{contributions} of this work are threefold: (1) A novel
support-enhanced granular-jamming gripper is designed for tendon-driven
continuum manipulators. It combines a compliant membrane and granular filler
with an internal support rod that promotes lateral enclosure during contact
and establishes a direct load-transfer path after jamming. The supported
configuration succeeds in 10/10 lift-and-hold trials, compared with 6/10
without the support rod (Fig.~\ref{fig:configuration_screening}C), while its
geometry-dependent grasp boundaries are further characterized under contact
offsets and different object shapes (Figs.~\ref{fig:contact_tolerance}
and~\ref{fig:shapes}). (2) A new autonomous continuum-manipulation system is
developed around the proposed gripper. It integrates visual feedback, tendon
actuation, and task-level pneumatic control, with learned control governing
target reaching and passive mechanics governing grasp formation. The physical
system demonstrates the complete
reach--grasp--lift--transfer--release sequence
(Fig.~\ref{fig:pick_drop}). (3) A new continuum-specific reaching formulation
is developed using reinforcement learning. It combines a base-fixed relative
target representation, a four-frame observation history, absolute
tendon-length actions, and targeted domain simulation to train the
policy using proximal policy optimization (PPO), as summarized in
Fig.~\ref{fig:rl_pipeline}. Trained entirely in simulation, the policy is
deployed without fine-tuning on physical data and achieves a terminal
root-mean-square error (RMSE) of $4.23~\mathrm{cm}$ over 16 physical rollouts,
compared with $2.51~\mathrm{cm}$ in simulation.

\section{Related Work}

This section reviews three areas relevant to the proposed system: continuum-manipulator modeling and control, mechanically adaptive grippers, and learning-based sim-to-real transfer. The discussion focuses on how these approaches reduce or accommodate uncertainty during reaching and grasping.

\subsection{Continuum-Manipulator Modeling and Control}

Analytical models relate continuum deformation to actuator inputs. Jones and
Walker derived kinematic relationships for multisection continuum
manipulators~\cite{Jones2006Kinematics}. Variable-curvature models represent
deformation beyond the constant-curvature assumption
~\cite{Mahl2014VariableCurvature}, while Rucker and Webster modeled
tendon-driven continuum robots under general loading and tendon-routing
conditions~\cite{Rucker2011Statics}. Although these models support analysis
and control, their accuracy depends on mechanical parameters and loading
conditions that can vary during physical operation.

Feedback methods reduce this dependence by correcting motion from measured
states. Visual pose estimation provides continuum-configuration or tip-motion
information~\cite{Reiter2011VisualPose}, and model-less feedback control uses
such measurements without requiring a complete analytical model
~\cite{Yip2014ModelLess}. Tendon-based stiffening and variable-stiffness
structures further improve stability
~\cite{Shiva2016TendonStiffening,Manti2016Stiffening}. These methods primarily
reduce uncertainty through body modeling, feedback control, or stiffness
regulation. Our method complements them by introducing mechanical
adaptability at the end effector, preserving continuum-body compliance while
enlarging the set of contact configurations from which stable enclosure can
form.

\subsection{Mechanically Adaptive Grippers}

Mechanically adaptive grippers increase object contact through deformation,
wrapping, or enclosure. Spiral and deformable-mesh mechanisms surround objects
without conventional finger closure
~\cite{Wang2025SpiRobs,Chen2023WebGripper}. Filament-weaving and tendril-like
kirigami mechanisms generate capture through structural deformation
~\cite{Kang2023DynamicWeaving,Hong2023Tendril}, while kirigami shells conform
through geometric reconfiguration~\cite{Yang2021KirigamiShells}. Their
deformation paths, structural dimensions, and actuation requirements,
however, can complicate integration with a slender and load-sensitive
continuum manipulator.

Granular-jamming grippers instead conform through particle rearrangement
inside a flexible membrane and stiffen under vacuum
~\cite{Brown2010Universal}. Pneumatic variants provide additional contact and
release behaviors~\cite{Amend2012PositivePressure}, while membrane, particle,
and filling properties influence jamming performance
~\cite{Fitzgerald2020JammingReview}. Our design retains passive conformity and
pneumatic locking but introduces an internal support for operation at a
bending continuum tip. Compared with an unsupported configuration, the
support promotes lateral enclosure and provides a more direct load-transfer
path after jamming, making the gripper better suited to
continuum-manipulator integration.

\subsection{Learning-Based Control and Sim-to-Real Transfer}

Learned forward and inverse models can represent the nonlinear response of
continuum manipulators~\cite{Thuruthel2017LearningDynamics}. Reinforcement
learning instead optimizes a control policy through task interaction, but a
policy trained in simulation may depend on dynamics that do not hold on
physical hardware.

Domain randomization improves transfer by exposing a policy to varied
simulation conditions. Tobin et al.\ randomized visual properties
~\cite{Tobin2017DomainRandomization}, whereas Peng et al.\ randomized physical
parameters~\cite{Peng2018DynamicsRandomization}. For a tendon-driven continuum
manipulator, initial posture, target position, effective stiffness, and
actuation delay directly affect reaching behavior. Our formulation combines
these continuum-specific randomizations with a four-frame observation history
and absolute tendon-length commands. This combination allows a standard PPO policy to respond to recent continuum motion while keeping the spatial and action representations approximately aligned between simulation and zero-shot deployment.

\section{System Overview}

This section describes the system architecture and autonomous task sequence. During operation, visual feedback guides tendon actuation toward the object, continued contact reshapes the unjammed gripper, and vacuum then locks the resulting configuration for lifting and transfer.

\subsection{System Architecture}

\begin{figure}[!t]
  \centering
  \includeorplaceholder{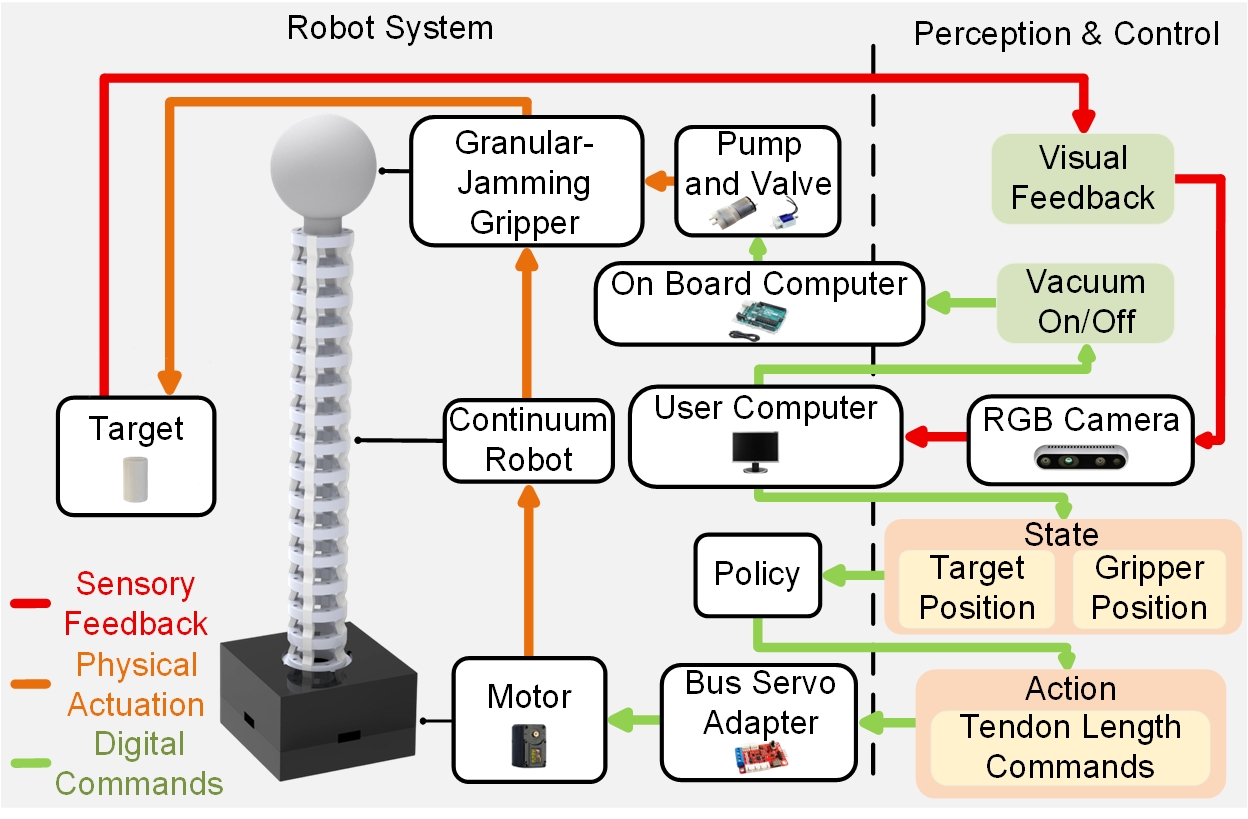}
    {0.98\columnwidth}{%
    Upload the system architecture as
    figures/fig2\_system\_architecture.jpg.}
  \caption{Architecture of the autonomous grasping system. Visual feedback
    provides the target and gripper states to the reaching policy, which outputs
    a four-dimensional absolute tendon-length command. The task layer coordinates
    pneumatic jamming and release, while the gripper forms and retains contact
    through mechanical deformation. This functional separation enables autonomous
    operation without requiring the learned policy to model granular contact
    mechanics.}
  \label{fig:system_architecture}
\end{figure}

As shown in Fig.~\ref{fig:system_architecture}, the system comprises a
tendon-driven continuum manipulator, a distal granular-jamming gripper, tendon
and pneumatic actuation units, an RGB camera, and a control computer. Four
independently regulated tendons determine the continuum body's bending
direction and magnitude. The camera provides the target and gripper positions
to the control computer, which generates tendon commands, while a pump and
valves switch the gripper between ambient-pressure and vacuum states.

The learned policy controls tendon-based reaching, the task layer coordinates
pneumatic switching, and the gripper mechanically forms and retains contact.

\subsection{Autonomous Task Sequence}

The task lifts an object from the table and releases it over a designated
region rather than at a unique pose. Under visual feedback, the unjammed
gripper approaches the object. Once the reaching condition is met, continued
approach develops contact and enclosure before the task layer applies vacuum.
The manipulator then lifts, transfers, and releases the object over the
destination.

\begin{figure}[!t]
  \centering
  \includeorplaceholder{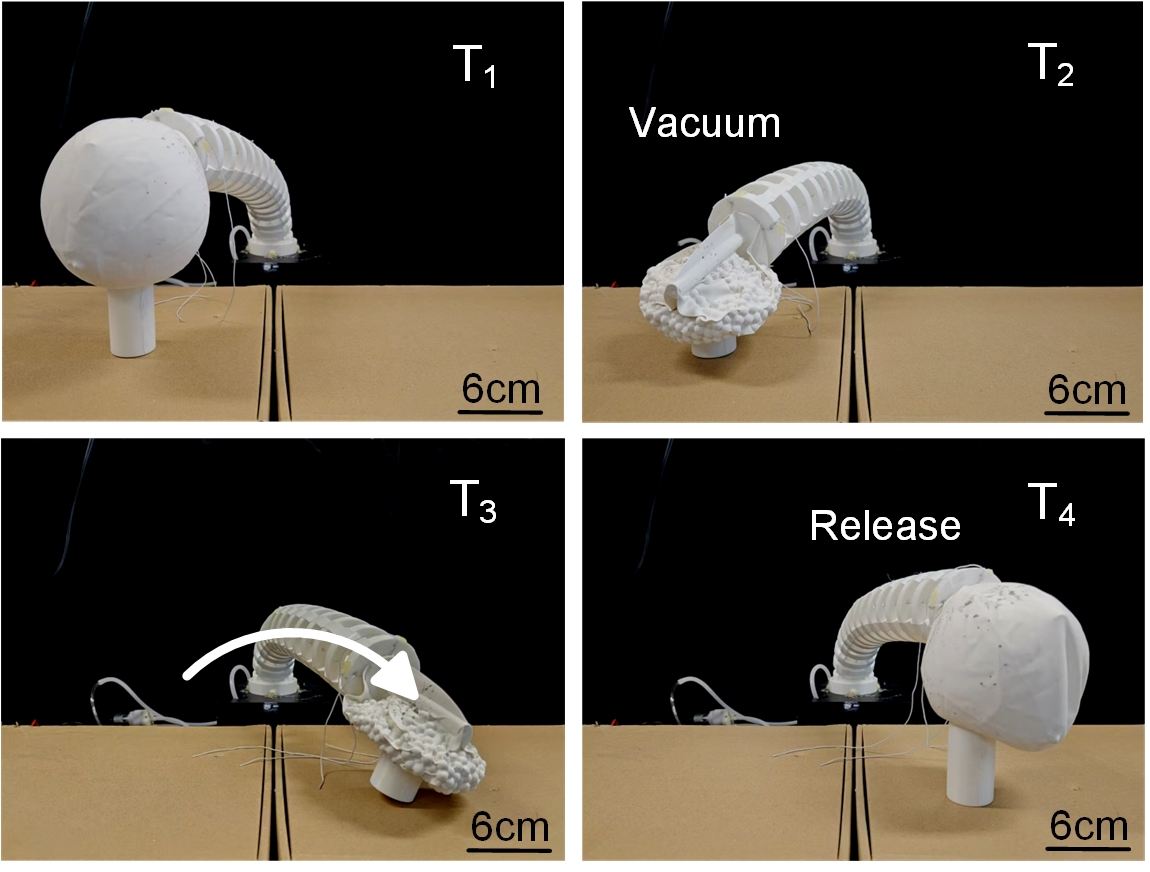}
    {0.98\columnwidth}{%
    Upload the real-system task sequence as
    figures/fig3\_task\_sequence.jpg.}
  \caption{Autonomous task sequence comprising policy-controlled approach,
    continued contact and enclosure, vacuum-induced jamming, lifting, transfer,
    and release. Only continuum reaching is learned; enclosure develops through
    passive mechanical deformation, while pneumatic switching is coordinated by
    the task layer. The sequence shows how learned reaching and mechanically
    adaptive grasp formation jointly produce the complete autonomous operation.}
  \label{fig:task_sequence}
\end{figure}

As summarized in Fig.~\ref{fig:task_sequence}, the policy need only reach a
finite region suitable for contact rather than a single exact grasp pose;
passive mechanics form the grasp, while the task layer controls jamming and
release.

\section{The Design of Granular-Jamming Gripper}
The gripper is designed to remain compliant during contact while providing stable force transmission after jamming. We first explain the mechanical role of the internal support rod and then evaluate the membrane--particle combination, filling ratio, and support configuration used in the subsequent experiments.

\subsection{Structure and Principle}

\begin{figure}[!t]
  \centering
  \includeorplaceholder{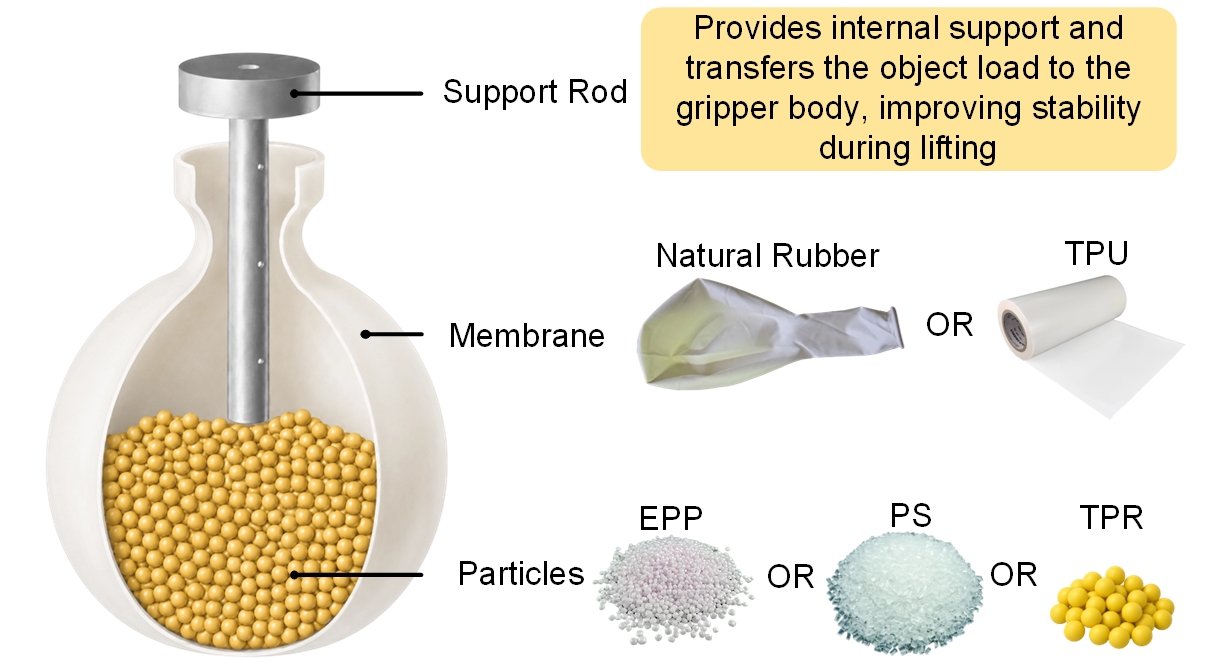}{0.98\columnwidth}{%
    Upload the gripper structure and candidate materials as
    figures/fig4\_gripper\_structure.jpg.}
  \caption{Structure of the support-enhanced granular-jamming gripper. The
    flexible membrane and granular filler provide contact conformity, while the
    internal support promotes lateral enclosure, transmits compressive force
    during contact, and provides a direct load path after jamming. The structure
    therefore combines a compliant contact surface with internal force
    transmission suitable for a bending continuum tip.}
  \label{fig:gripper_structure}
\end{figure}

The gripper comprises a flexible membrane, granular filler, an internal
support rod, and a pneumatic interface, as illustrated in
Fig.~\ref{fig:gripper_structure}. The membrane forms a deformable outer surface
and confines the particles. Before evacuation, the particles move relative to
one another and allow external contact to reshape the gripper.

The support rod performs three coupled functions. First, it provides a local
internal boundary that supports lateral enclosure. Second, it transmits
compressive force from the continuum tip into the
membrane--particle--object contact. Continued approach can therefore
redistribute the particles and deepen the enclosure. Third, after jamming,
the rod provides a more direct load path from the object to the continuum tip,
reducing reliance on membrane tension alone.

At ambient pressure, the membrane continues to deform after contact as the
particles rearrange around the object. Evacuation increases interparticle
normal forces and frictional constraints, suppressing relative particle motion
and locking the resulting configuration. Load bearing therefore does not
begin at first contact. Limited continued compression is used to establish a
more complete enclosure before vacuum-induced jamming.

\begin{figure}[!t]
  \centering
  \includeorplaceholder{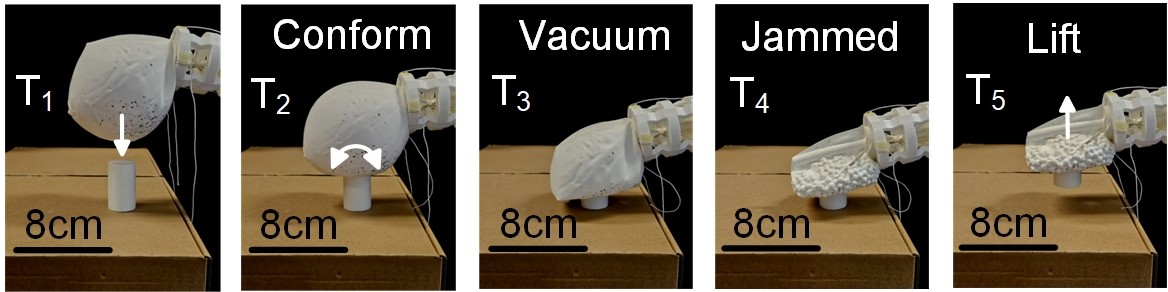}{0.98\columnwidth}{%
    Upload the gripper operating sequence as
    figures/fig5\_gripper\_operation.jpg.}
  \caption{Operating sequence of the granular-jamming gripper. Contact and
    continued compression first redistribute the particles around the object,
    after which evacuation locks the adapted configuration. The pressure
    transition converts contact-generated conformity into a load-bearing grasp
    supported by the jammed particles and internal rod.}
  \label{fig:gripper_operation}
\end{figure}

The sequence in Fig.~\ref{fig:gripper_operation} shows the two mechanical
states used by the system. The unjammed state provides conformity during
contact, whereas the jammed state provides retention during lifting and
transfer.

\subsection{Materials Configuration}
\label{subsec:configuration_selection}

The configuration study addresses three questions: which membrane--particle
combination provides reliable grasping with low distal load, which filling
ratio balances conformity and retention, and whether the support rod improves
grasping under otherwise matched conditions. Each condition is evaluated in
ten trials using a common test object, evacuation condition, operating
procedure, and success criterion. A trial is successful when the object is
lifted clear of the table and retained through the lifting phase.

\begin{figure*}[!t]
  \centering
  \includeorplaceholder{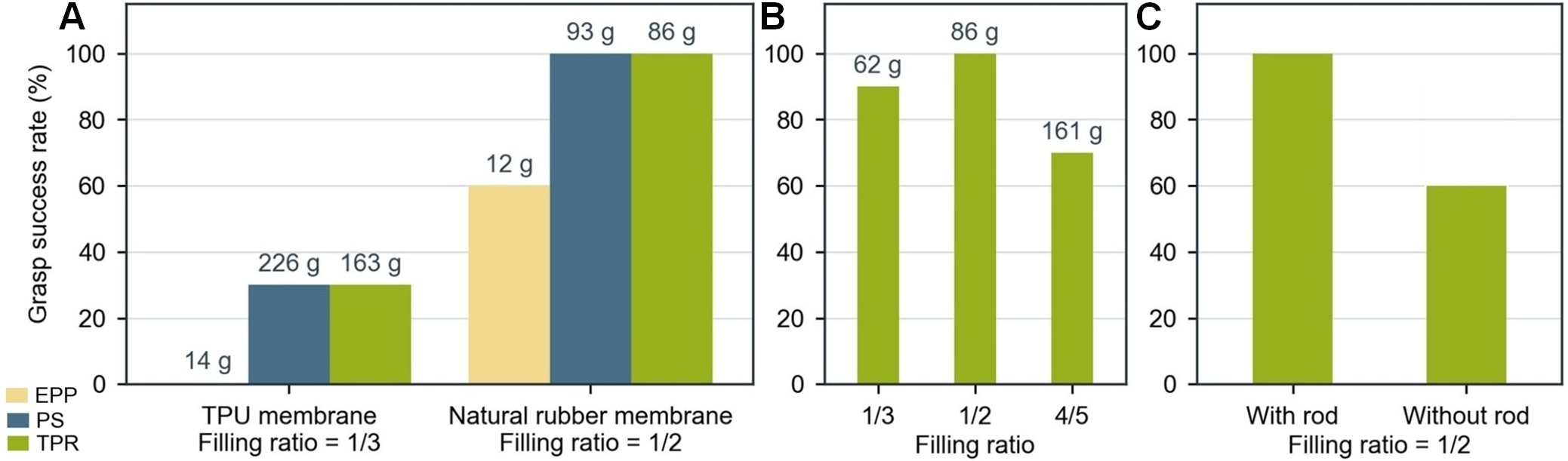}
    {0.94\textwidth}{%
    Upload the combined configuration comparison as
    figures/fig6\_configuration\_screening.jpg.}
  \caption{Configuration selection and support-rod evaluation. (A) Grasp
    success rates for combinations of natural-rubber or thermoplastic-polyurethane
    membranes with expanded-polypropylene, polystyrene, or thermoplastic-rubber
    particles. (B) Success rates at nominal filling ratios of $1/3$, $1/2$, and
    $4/5$ using the natural-rubber membrane and TPR spheres. (C) The selected
    configuration succeeds in 10/10 trials with the support rod and 6/10 trials
    without it. Each condition comprises ten trials. These results select the
    natural-rubber membrane, TPR spheres, and 50\% filling ratio, and demonstrate
    that the internal support increases the measured grasp success rate from
    60\% to 100\% under the tested condition.}
  \label{fig:configuration_screening}
\end{figure*}

\textbf{Membrane and Particle Selection:}
We compare natural-rubber and thermoplastic polyurethane (TPU) membranes with
expanded-polypropylene (EPP) granules, polystyrene (PS) particles, and
thermoplastic-rubber (TPR) spheres. Under the present experimental conditions,
the natural-rubber membrane outperforms the TPU membrane. With the
natural-rubber membrane, both PS particles and TPR spheres achieve a 100\%
grasp success rate. We select the TPR spheres because they produce a lower
total gripper mass while retaining the same measured success rate, thereby
reducing the distal load on the compliant continuum manipulator.

\textbf{Filling-Ratio Selection:}
The nominal filling ratio is
$\eta=V_{\mathrm{p}}/V_{\mathrm{m}}$, where $V_{\mathrm{p}}$ is the
loose-particle volume and $V_{\mathrm{m}}$ is the membrane interior volume in
its natural, uncompressed state. As shown in
Fig.~\ref{fig:configuration_screening}B, the natural-rubber membrane with TPR
spheres achieves success rates of 90\%, 100\%, and 70\% at filling ratios of
$1/3$, $1/2$, and $4/5$, respectively. A low filling ratio provides more free
deformation space but fewer particles for load bearing. An excessively high
ratio restricts conformity and increases distal load. The final gripper
therefore uses a 50\% nominal filling ratio.

\textbf{Support-Rod Evaluation:}
Using the selected membrane--particle combination and filling ratio, we compare
otherwise matched grippers with and without the support rod. Both
configurations use the same object, approach posture, evacuation condition,
and success criterion. The supported gripper succeeds in 10/10 trials, whereas
the rod-free gripper succeeds in 6/10 trials, as shown in
Fig.~\ref{fig:configuration_screening}C.

Without internal support, the gripper tends to hang downward as the continuum
body bends laterally. This reduces the membrane area presented to the object.
The unsupported distal body is also more susceptible to relative motion during
continuum oscillation, which can disturb retention after lifting. The support
rod maintains a more useful contact geometry, transmits compressive force to
promote enclosure, and provides a direct load-transfer path after jamming.
Because these functions are mechanically coupled, the experiment evaluates
their combined contribution rather than isolating them individually.

The configuration study therefore establishes the gripper used in the
remaining experiments: a natural-rubber membrane, TPR spheres, a 50\% nominal
filling ratio, and an internal support rod.

\section{The Characterization of the Gripper}

To isolate the gripper mechanics from the learned reaching controller, continuum motion was manually controlled or fixed at the corresponding task stage. Grasping performance was evaluated under contact-position offsets and across different object geometries, with ten trials per condition using the common procedure and success criterion defined in Section~\ref{subsec:configuration_selection}.

\subsection{Grasping Under Contact-Position Offset}

To evaluate imperfect alignment, we repeatedly grasp a cube at discrete
lateral and longitudinal offsets. In the repressurized state, the gripper has
an approximately spherical profile with a nominal diameter of
$10~\mathrm{cm}$. The origin $O$ is the orthogonal projection of the sphere's
geometric center onto the table. The signed displacements of the cube center
from $O$ are denoted by $\Delta x$ and $\Delta y$. This origin provides a
consistent geometric reference and is not assumed to be the optimum grasp
point.

\begin{figure}[!t]
  \centering
  \includeorplaceholder{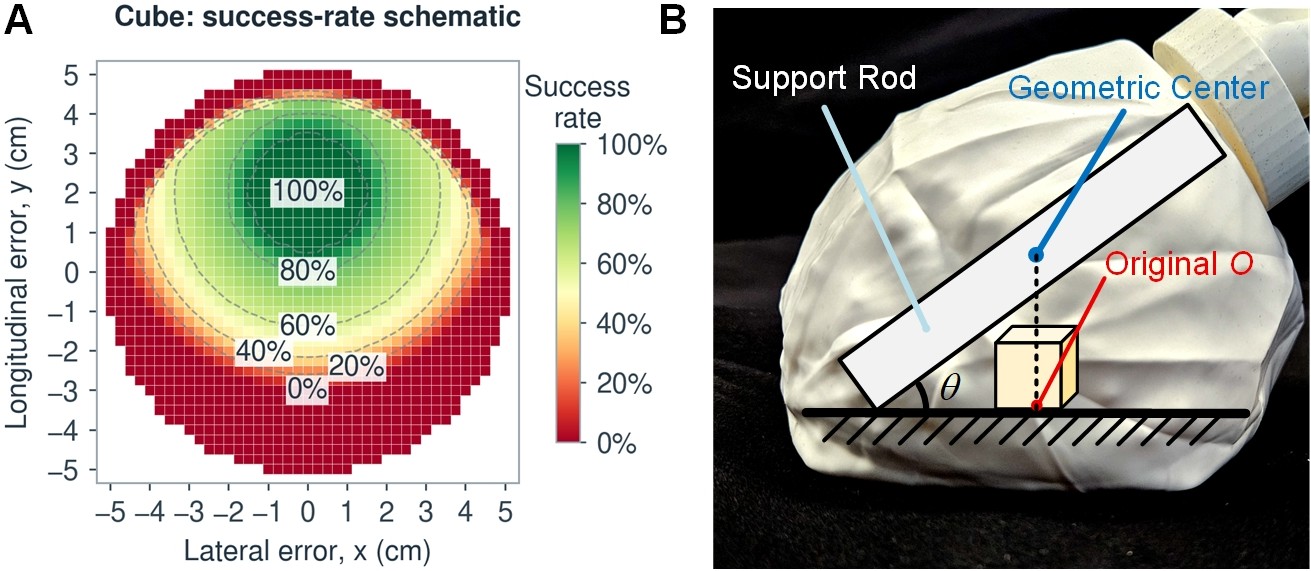}
    {0.98\columnwidth}{%
    Upload the contact-offset map and enclosure-geometry schematic as
    figures/fig7\_contact\_tolerance\_geometry.jpg.}
  \caption{Geometry-dependent grasping under contact-position offset. (A) Each
    marked position is tested ten times, and the continuous field is interpolated
    from the discrete measurements only to visualize the spatial trend. (B) The
    membrane, support rod, object, and table jointly form a stable enclosure under
    favorable contact configurations. The shifted high-success region shows that
    the gripper enlarges a geometry-dependent set of graspable configurations
    rather than providing a uniform, isotropic position-tolerance region.}
  \label{fig:contact_tolerance}
\end{figure}

Figure~\ref{fig:contact_tolerance}A visualizes the measured spatial trend. The
continuous color field and contours do not represent measurements at every
location. Figure~\ref{fig:contact_tolerance}B illustrates how the membrane,
support rod, object, and table form a joint enclosure under the current
support-rod orientation.

The graspable region is not a uniform circular area centered at $O$. In the
present configuration, the support rod approaches the object with an
approximately $30^\circ$--$40^\circ$ downward inclination relative to the
table. Objects toward its distal side more readily enter an enclosure formed
jointly by the membrane, rod, and table. This shifts the high-success region
away from the geometric center. Within these favorable configurations, the
support rod transmits continued approach force into the contact and helps the
membrane and particles develop a deeper enclosure.

The highest measured success rate is 100\%. The condition near $O$ yields
80\%, and selected lateral-offset conditions reach 90\%. The success rate
falls to approximately 50\% or below where the membrane, rod, object, and
table cannot establish a stable joint constraint. The gripper therefore
enlarges the set of contact configurations that permit enclosure rather than
providing a large, isotropic tolerance region. Because this set depends on
support-rod posture and environmental constraint, other approach orientations
may shift its location. This effect remains to be evaluated experimentally.

\subsection{Grasping Objects of Different Shapes}

We next evaluate whether the selected gripper can form a load-bearing
enclosure around different object geometries. Cylinders, cubes, spheres,
hexagonal prisms, and triangular pyramids are each tested in ten trials using
the common experimental protocol.

\begin{figure*}[!t]
  \centering
  \includeorplaceholder{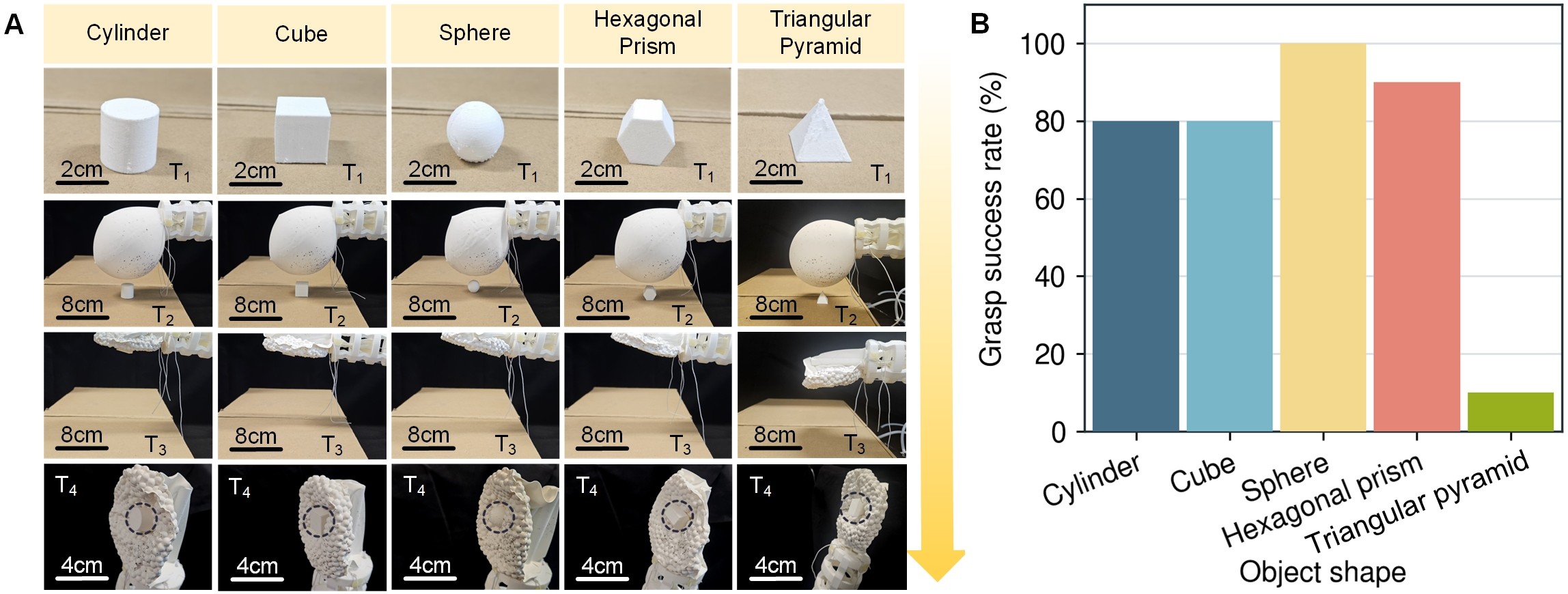}{0.96\textwidth}{%
    Upload the combined horizontal object-shape figure as
    figures/fig8\_shape\_results.jpg.}
  \caption{Grasping performance across object geometries. (A) Representative
    sequences for a cylinder, cube, sphere, hexagonal prism, and triangular pyramid.
    (B) Their success rates are 80\%, 80\%, 100\%, 90\%, and 10\%, respectively,
    over ten trials per shape. The results demonstrate adaptation to several
    geometries while identifying sharp-edged objects with limited stable contact
    as a limitation of the current design.}
  \label{fig:shapes}
\end{figure*}

As shown in Fig.~\ref{fig:shapes}, the sphere is grasped in 10/10 trials
(100\%), the hexagonal prism in 9/10 trials (90\%), the cylinder and cube each
in 8/10 trials (80\%), and the triangular pyramid in only 1/10 trials (10\%). Although its success rate is low, the successful trial shows that the current
gripper can still form a load-bearing enclosure around this challenging
geometry.

The continuous surfaces of the sphere and cylinder allow the membrane to
spread around the object and establish distributed enclosure. Although the
cube and hexagonal prism contain planar faces and edges, they still provide
sufficient contact area and geometric constraint for a load-bearing
configuration. In contrast, the sharp edges and small stable contact region of
the triangular pyramid promote local membrane folding and allow slip or rotation
before sufficient enclosure develops.

Together, the contact-offset and object-shape experiments show that mechanical
compliance does not provide unrestricted grasping tolerance. Instead, it
enlarges the set of relative configurations in which the membrane, particles,
support rod, and environment can form a load-bearing enclosure.

\section{Continuum Modeling and Reinforcement-Learning-Based Reaching}

The reaching controller is designed to bring the gripper into a state suitable for contact rather than to control grasp formation or pneumatic switching. This section formulates the reaching task using a nominal PCC model, a history-based observation, absolute tendon-length actions, targeted domain simulation, and PPO training for zero-shot physical deployment.

\subsection{Continuum Modeling}

We model the continuum using the piecewise-constant-curvature (PCC) modeling
~\cite{Webster2010ConstantCurvature}. For a segment of length $L$, the
curvature $\kappa$ and bending-plane angle $\phi$ are assumed to remain
constant along the arc length. The centerline position at $s\in[0,L]$,
expressed in the segment base frame, is
\begin{equation}
\bm{p}(s)=
\begin{cases}
\begin{bmatrix}
\dfrac{1-\cos(\kappa s)}{\kappa}\cos\phi \\
\dfrac{1-\cos(\kappa s)}{\kappa}\sin\phi \\
\dfrac{\sin(\kappa s)}{\kappa}
\end{bmatrix}, & \kappa\neq 0, \\[5mm]
\begin{bmatrix}0&0&s\end{bmatrix}^{\!\mathsf T}, & \kappa=0.
\end{cases}
\label{eq:pcc_position}
\end{equation}

The four tendon lengths jointly determine the nominal curvature and bending
direction. The PCC model provides a low-dimensional geometric basis for
policy training rather than an exact representation of the physical
continuum dynamics. In particular, it does not fully reproduce tendon
friction, hysteresis, material variation, actuation delay, external loading,
or the coupled deformation observed on the physical system. These
discrepancies motivate the use of observation history and targeted
randomization.

\subsection{Policy Observation and Action}

We define a base-fixed task frame $\mathcal{F}_{B}$ for reaching, with its
origin at the center of the continuum-manipulator base and its axes fixed to
the base. Camera-detected gripper and target positions are transformed into
this frame, and all position quantities below are expressed in
$\mathcal{F}_{B}$. The same frame convention is used in simulation and
physical deployment.

At time $t$, the gripper-center and target positions are denoted by
$\bm{p}_{g,t}$ and $\bm{p}^{*}_{t}$, respectively. The relative target
displacement is
$\Delta\bm{p}_{t}=\bm{p}^{*}_{t}-\bm{p}_{g,t}\in\mathbb{R}^{3}$,
which specifies the required motion direction and magnitude. Its norm
$d_t=\|\Delta\bm{p}_{t}\|_2$ defines the target distance used in the reward.
Each observation contains the gripper position
$\bm{p}_{g,t}\in\mathbb{R}^{3}$, relative target displacement
$\Delta\bm{p}_{t}\in\mathbb{R}^{3}$, and current tendon state
$\bm{\ell}_t\in\mathbb{R}^{4}$. Their concatenation is
$\bm{o}_t=\operatorname{col}
(\bm{p}_{g,t},\Delta\bm{p}_{t},\bm{\ell}_t)
\in\mathbb{R}^{10}$. Stacking four consecutive observations yields
$\bm{s}_t=\operatorname{col}
(\bm{o}_t,\bm{o}_{t-1},\bm{o}_{t-2},\bm{o}_{t-3})
\in\mathbb{R}^{40}$. This history provides short-term motion context
associated with hysteresis, stiffness variation, and actuation delay.

The policy outputs an absolute tendon-length command
$\bm{a}_t=\bm{\ell}^{\mathrm{cmd}}_t\in\mathbb{R}^{4}$. In the nominal model, tendon lengths define a nominal posture and correspond directly to the position-controlled tendon interface. The policy therefore outputs four bounded absolute tendon-length targets. With delayed feedback, errors in incremental commands can accumulate, causing command drift or, without saturation, targets outside the allowable range. However, friction, hysteresis, stiffness variation, and external loading can make the same targets produce different postures. Visual feedback and observation history help the policy respond to these deviations.

\subsection{Reward Design}

The reward is designed to distinguish approaching the target from reaching
and maintaining a contactable state. It is written compactly as
$r_t=r_t^{\mathrm{approach}}+r_t^{\mathrm{reach}}
+r_t^{\mathrm{slow}}+r_t^{\mathrm{hold}}$.

The approach term rewards a reduction in target distance between consecutive
steps and therefore provides a dense learning signal before the target region
is reached. The reach term identifies entry into the target neighborhood.
Distance alone is insufficient because a compliant manipulator may pass
through this region with substantial residual motion. The slow-motion term
therefore discourages aggressive motion near the target. Finally, the hold
term rewards maintaining the reached state, preventing a transient pass
through the target region from being treated as successful reaching. Together,
these terms train the policy to approach, slow down, and remain near the
target long enough for the task layer and compliant gripper to initiate
contact and enclosure.

\subsection{domain simulation}

\begin{figure}[!t]
  \centering
  \includeorplaceholder{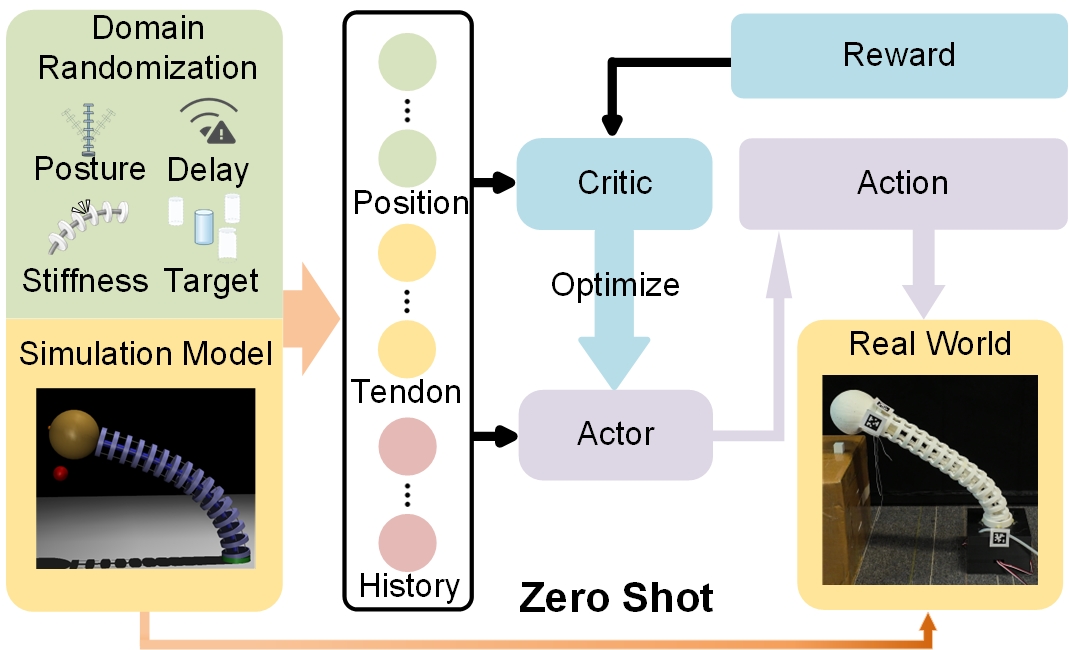}{0.98\columnwidth}{%
    Upload the reinforcement-learning diagram as
    figures/fig9\_rl\_pipeline.jpg.}
 \caption{Randomized training and zero-shot physical deployment of the reaching
    policy. In a task frame fixed to the continuum-manipulator base, the policy
    receives four consecutive observations of the gripper position, relative
    target displacement, and tendon state, and outputs a four-dimensional absolute
    tendon-length command. Randomizing the initial posture, target position,
    stiffness, and actuation delay exposes the policy to continuum-specific
    variations. The formulation preserves consistent state and action meanings
    between simulation and the physical system without requiring the policy to
    learn contact deformation or pneumatic control.}
  \label{fig:rl_pipeline}
\end{figure}

The simulation randomizes the initial posture, target position, effective
stiffness, and actuation delay, with each variation representing a source of
physical deployment uncertainty. Initial-posture randomization prevents the
policy from relying on a single reset configuration, while target-position
randomization trains goal-conditioned reaching instead of one fixed tendon
trajectory. Effective stiffness is varied to represent changes in material
properties, assembly, and loading that alter the tendon-length-to-curvature
relationship. Actuation-delay randomization accounts for delays introduced by communication, actuator response, and tendon transmission. Together with the four-frame observation history, these variations
encourage the policy to respond to the observed continuum motion rather than
rely on one fixed tendon-to-position mapping or action-response timing. The
randomized training and deployment pipeline is summarized in
Fig.~\ref{fig:rl_pipeline}.

\subsection{PPO Training and Deployment}

We train the reaching controller using PPO
~\cite{Schulman2017PPO}. The policy maps the four-frame state to a distribution
over absolute tendon-length commands, while a value network estimates the
expected return during training. Both networks are updated using rollouts
collected under the randomized simulation conditions described above.

During physical deployment, simulated positions are replaced by camera
measurements transformed into the same base-fixed task frame
$\mathcal{F}_{B}$. The state structure and absolute tendon-length action remain
unchanged between simulation and the physical system. Only the trained policy
is retained, and no gradient update or physical-system fine-tuning is
performed. Grasp formation and pneumatic switching remain outside the learned
controller.

The policy is evaluated in 16 physical rollouts across two target positions.
Its terminal root-mean-square error (RMSE) is $4.23~\mathrm{cm}$ on the physical system, compared with
$2.51~\mathrm{cm}$ in simulation. These results characterize the zero-shot
transfer of the reaching policy; the gripper's mechanical grasp range is
evaluated separately in the contact-offset experiments.

\begin{figure*}[!t]
  \centering
  \includeorplaceholder{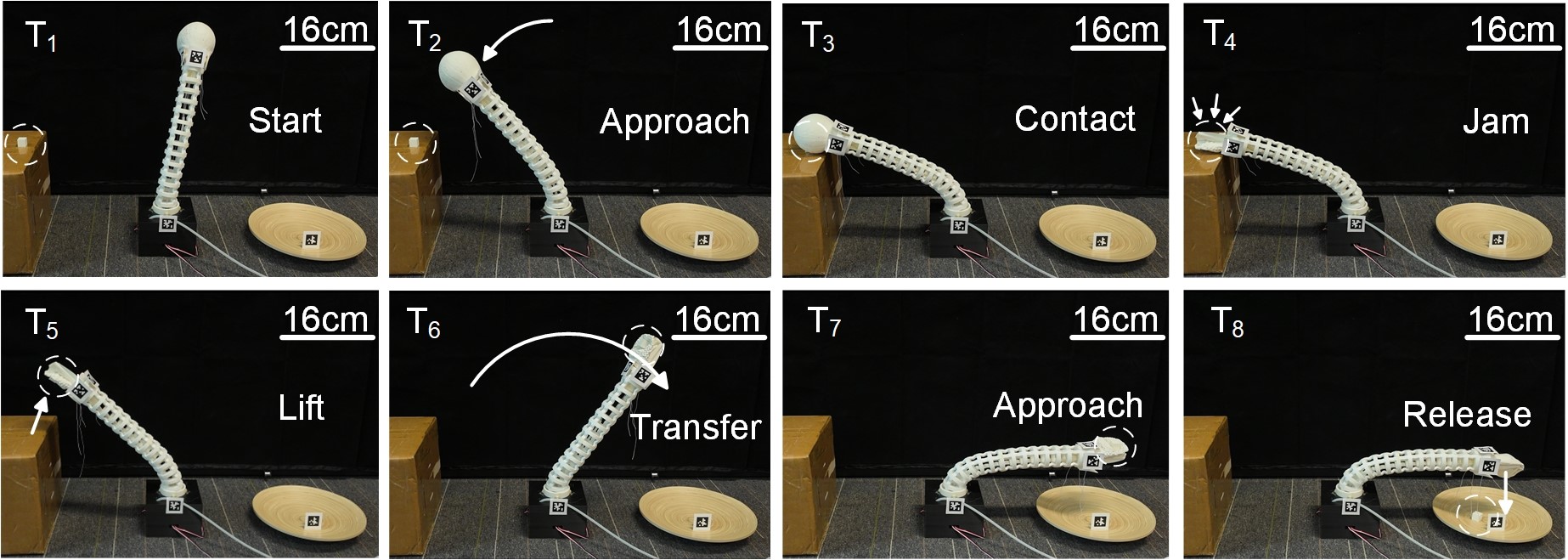}
    {0.94\textwidth}{%
    Upload the physical reach--grasp--transfer--release sequence as
    figures/fig11\_pick\_release\_sequence.jpg.}
  \caption{Autonomous grasp-and-release sequence on the physical system:
    policy-controlled approach, contact and passive conformity, task-level vacuum
    activation, lifting, transfer, and venting over the destination. The
    completed sequence demonstrates that zero-shot continuum reaching, the
    support-enhanced gripper, and pneumatic task control can operate together
    without requiring the policy to model or control granular contact
    deformation.}
  \label{fig:pick_drop}
\end{figure*}

\section{Autonomous Grasp-and-Release Experiments}

Finally, we evaluate whether the independently developed gripper and zero-shot reaching policy can operate together in a complete autonomous task. The experiment combines visual state estimation, learned tendon control, passive grasp formation, and task-level pneumatic switching to execute the reach--grasp--lift--transfer--release sequence.

Figure~\ref{fig:pick_drop} shows the gripper remaining unjammed during approach, deforming around the object during continued contact, retaining the object after evacuation, and releasing it over the destination.

The demonstration validates the functional integration of visual state
estimation, learned tendon control, passive contact adaptation, and pneumatic
switching. It shows that the support-enhanced gripper can be incorporated into
an autonomous continuum-manipulation system while preserving a clear
separation between learned reaching and mechanically adaptive grasp formation.

\section{Conclusion and Future Work}

This work presented a lightweight, support-enhanced granular-jamming gripper
for autonomous grasping with a tendon-driven continuum manipulator. Before
evacuation, the flexible membrane and mobile particles conform to the object
and accommodate imperfect contact. During contact, the internal support rod
promotes lateral enclosure and transmits compressive force from the continuum
tip. After evacuation, the jammed particles and support rod provide retention
and load transfer.

We systematically evaluated the membrane material, particle material, filling
ratio, and support structure. The selected configuration achieved 10/10
successful lift-and-hold trials with the support rod, compared with 6/10
trials without it. Contact-offset experiments showed that the gripper enlarges
a geometry-dependent set of successful contact configurations rather than
providing an isotropic position tolerance. Object-shape experiments further
demonstrated successful enclosure across several geometries while identifying
the triangular pyramid as a difficult case.

The gripper was further integrated with visual feedback, tendon actuation,
pneumatic control, and a history-based PPO reaching policy. The policy was
trained in a PCC-based simulation with randomized initial posture, target
position, stiffness, and actuation delay, and was deployed without
physical-system fine-tuning. The integrated system demonstrated the complete
autonomous reach--grasp--lift--transfer--release sequence. These results
establish that the support-enhanced gripper can provide mechanically adaptive
grasp formation within an autonomous continuum-manipulation system.

The current prototype still exhibits geometry-dependent graspability, and its
performance has mainly been evaluated under a limited set of approach
conditions and object geometries. The geometry of the internal support
structure has also not been fully optimized. Future work will optimize the
support structure, extend the graspable region to broader three-dimensional
position and orientation variations, and incorporate contact or pressure
sensing for more adaptive jamming and manipulation.

\FloatBarrier
\balance

\section*{ACKNOWLEDGEMENT}
This work was in part supported by the InnoHK initiative of the Innovation and Technology Commission of the Hong Kong Special Administrative Region Government via the Hong Kong Centre for Logistics Robotics. This work was also in part supported by the Chinese University of Hong Kong, Zhejiang University, and National Natural Science Foundation of China.

Generative AI tools were used to assist with grammar proofreading, and code debugging. All technical content and final manuscript decisions were developed and verified by the authors.

\bibliographystyle{IEEEtran}
\bibliography{references}

\end{document}